\documentclass{article} 
\usepackage{iclr2019_conference,times}
\iclrfinaltrue

\usepackage{amsmath,amsfonts,bm}

\def\eqref#1{equation~\ref{#1}}

\def\1{\bm{1}}

\DeclareMathAlphabet{\mathsfit}{\encodingdefault}{\sfdefault}{m}{sl}
\SetMathAlphabet{\mathsfit}{bold}{\encodingdefault}{\sfdefault}{bx}{n}

\usepackage{hyperref}
\usepackage{url}
\usepackage{graphicx}

\title{AI-based evaluation of the SDGs: \\ 
The case of crop detection with earth observation data}

\author{Natalia Efremova  \thanks{Centre for Corporate Reputation and Future of Marketing Initiative, University of Oxford}\\
University of Oxford\\
Oxford, UK\\
\texttt{natalia.efremova2@sbs.ox.ac.uk} \\
\And
Dennis West \thanks{Centre for Corporate Reputation, University of Oxford}\\
University of Oxford\\
Oxford, UK\\
\texttt{dennis.west@gtc.ox.ac.uk} \\
\AND
Dmitry Zausaev \\
Deep Planet\\
Herwell, UK \\
\texttt{dmitry@deepplanet.ai}
}

\begin{document}

\maketitle

\begin{abstract}
The framework of the seventeen sustainable development goals is a challenge for developers and researchers applying artificial intelligence (AI).  AI and earth observations (EO) can provide reliable and disaggregated data for better monitoring of the sustainable development goals (SDGs). In this paper, we present an overview of SDG targets, which can be effectively measured with AI tools. We identify indicators with the most significant contribution from the AI and EO and describe an application of state-of-the-art machine learning models to one of the indicators. We describe an application of U-net with SE blocks for efficient segmentation of satellite imagery for crop detection. Finally, we demonstrate how AI can be more effectively applied in solutions directly contributing towards specific SDGs and propose further research on an AI-based evaluative infrastructure for SDGs. 
\end{abstract}

\section{Introduction}
The framework of the seventeen sustainable development goals (SDGs) is a challenge for developers and researchers applying artificial intelligence (AI). The 169 targets are measured by 232 indicators each of which require a dedicated "evaluative infrastructure" \citep{kornberger}.  Statistical standard-setting within the United Nations is technically and politically complex. The estimated direct cost of measuring all SDGs are over \$US 250 billion, excluding opportunity costs \citep{jerven}.  Many indicators are at risk of elimination in the following assessment rounds by the technical commission of UN Statistics \citep{stat1}.  If the global community does not come up with generally accepted methodologies and if countries are unable to adopt them effectively, the SDGs cannot be monitored which is a repetition of the failures in the preceeding millennium development goals. \\
The United Nations already have a loose network of actors and processes that are related to AI \citep{un}.  We, therefore, argue that the primary purpose of the AI for Good movement is achieving the SDG target 17.19: building a systematic partnership to develop measurements of progress on sustainable development that complement GDP. Under the current framework, this target is primarily measured by the \$US value of all resources made available to strengthen statistical capacity in developing countries (SDG 17.19.1). AI for Good can contribute in three ways. First, we can help decreasing the cost of data collection and analysis. Second, we can help to enhance the capacity for measurement. This systematic approach allows, thirdly, to embed AI solutions within direct interventions more effectively.\\
One step towards this goal is using AI with earth observation data (EO). AI and EO can provide reliable and disaggregated data for better monitoring of the SDGs. Building on the existing work in relation to poverty and agricultural yields \citep{burke},  our project is in the context of yield prediction. Drawing on studies of calculative practices \citep{miller}, table \ref{role-evaluating} below shows four main roles of AI as part of an evaluative infrastructure, starting from a mapping role based on earth observation data.  Earth observation data is freely available and highly accurate. These data can further be combined on the socio-economic, organizational and institutional level whereby the roles of AI can be mediating, adjudicating, and ranking.\\ 
Table 2  lists non-exhaustively the SDGs indicators that can be addressed by AI and EO data. asterisk (*) indicates the SDGs, addressed by this project or those that potentially could be affected by the outcomes of the proposed approach. We distinguish between 1st generation and 2nd generation of AI-EO applications to SDGs (with more sophisticated AI applications). Some international bodies and working groups suggest the use satellite imaging data for several SDGs \citep{ppt1}, \citep{ppt2}, \citep{ppt3}.  Those proposals center around SDGs, whose indicators primarily use geographic data.  A further step is to use AI and EO in contexts where only small sample sizes are available or where states lack the capability to collect and analyze the data. Open source GIS and data analysis techniques allow us to evaluate progress towards the SDGs and strengthen accountability.\\
The UN classifies indicators into tier 1 whereby two criteria are met. First, a generally accepted methodology exists (methodology criteria). Second, this methodology is widely adopted around the world and states generate sufficient data  (adoption criteria). Tier 2 indicators do not meet either the methodology or adoption criteria. Tier 3 indicators fail to meet both criteria. Therefore, the most significant contribution of AI and EO can be made regarding tier 2 and 3 indicators. We also note that we have found some tier 1 indicators that are insufficiently measuring the intended target (ex. climate action targets 13.2 and 13b with indicators 13.2.1 and 13.b.1). As a result, more SDGs could be identified for improving tier 1 indicators through a systematic AI and EO review.\\
Tier 3 indicators would contribute most from the application of AI-based methods, therefore we consider an example of such an application in this paper. In the next sections, we will propose a model that can be used to evaluate indicator 2.4.1 "Proportion of agricultural area under productive and sustainable agriculture". This task could be decomposed into several sub-tasks, which could be solved using one machine learning method each. These sub-tasks include segmentation of satellite imagery to detect crops, which are currently growing in the region; detection of potentially more efficient crops in terms of scarce resource consumption (e.g. fresh water); estimation of the amount of nutrients in the soil; detection of soil moisture and salinity and crop yield prediction. The focus of this paper is on sub-task one (segmentation of satellite imagery for crops detection).

\begin{table}[t]
\caption{Role of AI in the evaluative infrastructure for SDGs}
\label{role-evaluating}
\begin{center}
\begin{tabular}{lll}
\multicolumn{1}{c}{\bf TYPE OF DATA}  &\multicolumn{1}{c}{\bf DATA ANALYSIS}&\multicolumn{1}{c}{\bf ROLE OF AI}
\\ \hline \\
Geographical            & Constructing global calculative space & Mapping\\
Socio-economic          & Identifying needs and vulnerabilities  & Mediating\\
Organisational             & Performance measurement & Adjudicating\\
Institutional 	& Rating and standardisation	& Ranking\\
\end{tabular}
\end{center}
\end{table}

\section{Methods}
\label{methods}
In the field of satellite imagery processing for agriculture, it is typically very difficult to obtain labelled datasets: data processing and labelling can be performed either by a professional agronomists or a farmer, who is familiar with particular crops. Either way, labelled data is rare or, in some cases, non-existent. Therefore, the approaches that are usually used in medical imaging with very little training data can be successfully applied to EO imagery segmentation. For crop identification task, we decided to use U-net \citep{unet}. U-net showed good results in the cases where there were very few annotated data samples available, since it relies on the strong use of data augmentation to use the available samples more efficiently.
\begin{figure}[h]
    \begin{center}
      \includegraphics[width=1\textwidth]{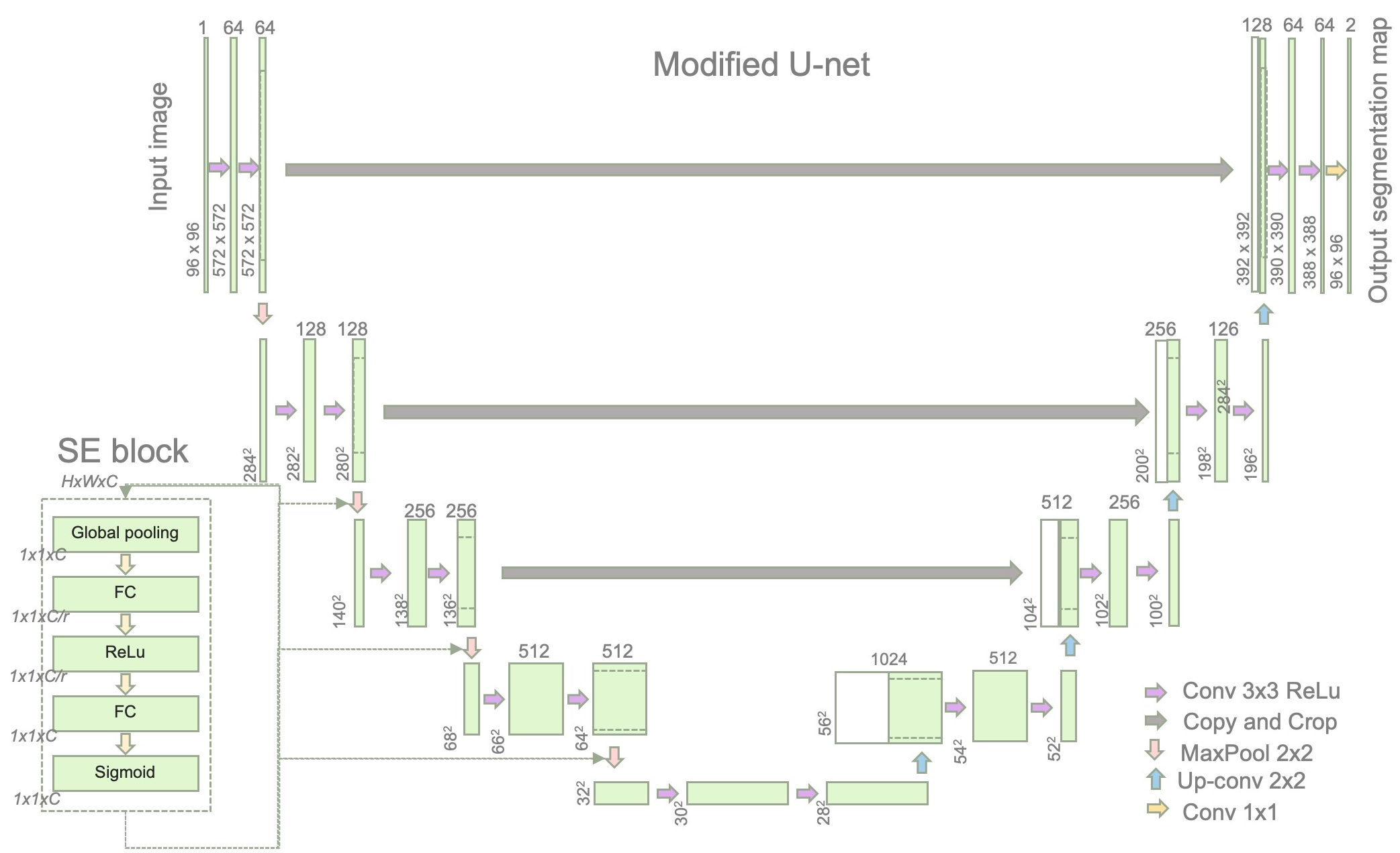}
        \caption{The architecture of the modified U-net model. Green boxes correspond to multi-channel feature maps. Different operations are denoted with coloured arrows. The structure of SE blocks is depicted in the bottom-left corner of the figure (all SE blocks share the same structure). The places where the SE blocks are inserted in the model are depicted with the dotted arrows.}
        \label{fig:unet}
    \end{center}
\end{figure}
We used two sources of satellite imagery: Sentinel-2 satellite and Google Earth. While Sentinel-2 is potentially more interesting for agricultural purposes, Google Earth data provides better ground resolution. As an example of the crops, we used grapes, since it has more available data for training. As a training dataset, we used one field in New South Wales region (Australia), taken at different timestamps and 12 vineyards in Napa Valley, California. As a test set, we have taken another field in New South Wales. Ground-truth labelling was performed in-house manually with \textit{kmz} polygons. \\
We have conducted a series of experiments to increase the performance of U-net on our task. We have tried a few traditional tools for increasing the accuracy of the model: residual blocks \citep{He2015} and squeeze-and-excitation (SE) blocks  \citep{se}. Addition of the residual blocks increased the performance of the model for up to 0,15\%. We assume that the reason for such an insignificant increase was the fact that basic U-net architecture is already well-optimised for extraction of the important features. Addition of squeeze-and-excitation (SE) blocks has significantly increased the quality of the model with neglectable increase in computational costs. We have added SE blocks into the encoding sections of our network as depicted in Fig.\ref{fig:unet}. It allowed us to increase the accuracy of the original architecture by over 4\%. 
\begin{figure}[h]
    \begin{center}
      \includegraphics[width=0.8\textwidth]{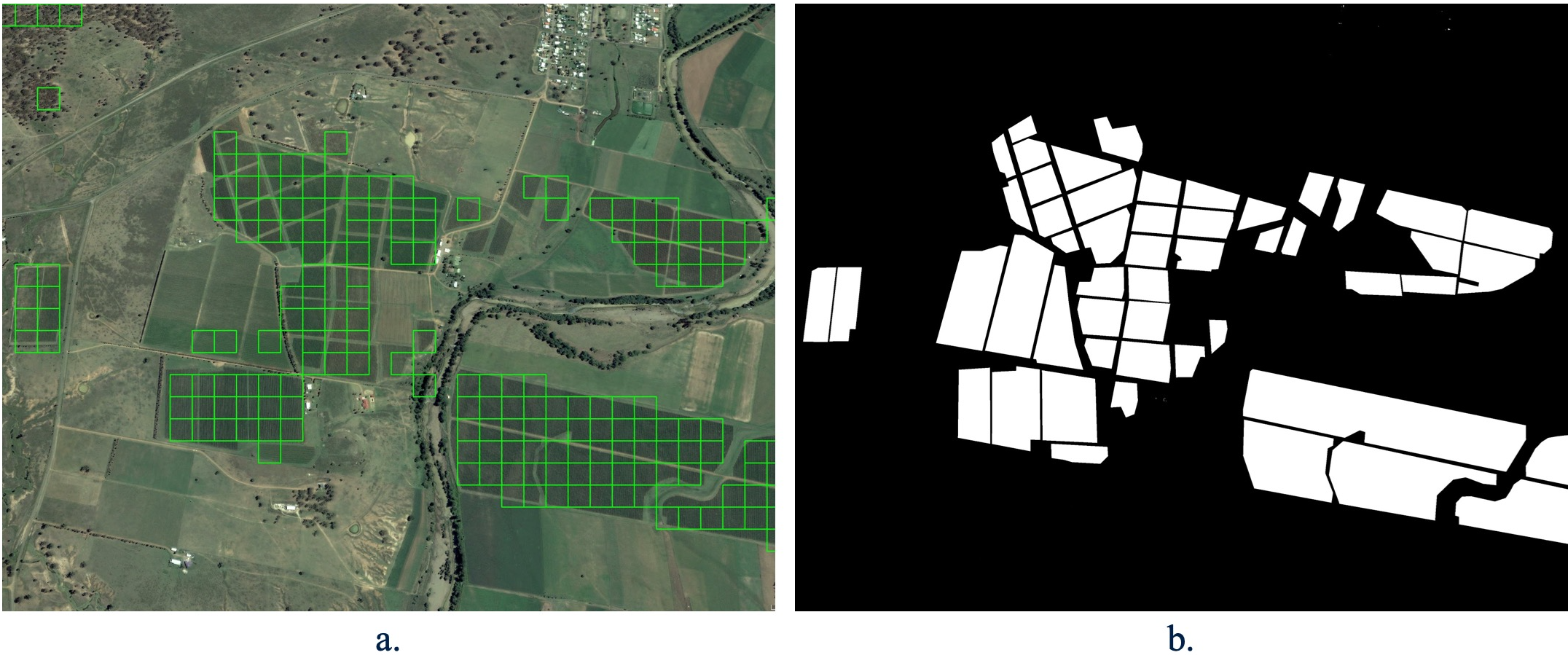}
        \caption{Segmentation results on the training set: (a) raw image; (b) generated segmentation mask (white: predicted polygons, black: background).}
        \label{fig:inp-out}
    \end{center}
\end{figure}

\begin{figure}[ht]
    \begin{centering}
            {\includegraphics[width=0.7\textwidth]{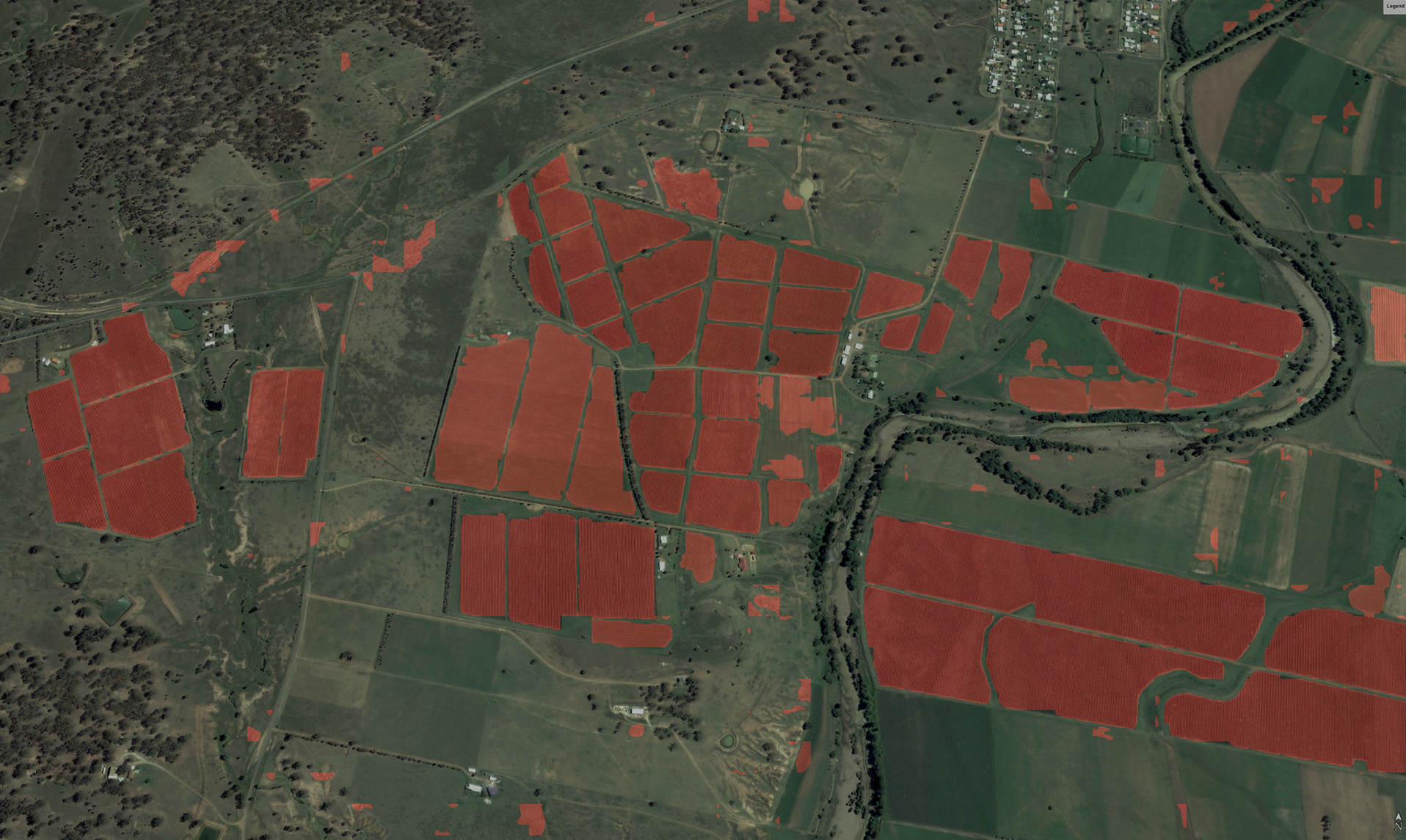}}
        \caption{Land cover segmentation overlay with ground truth segmentation.}
        \label{fig:seg}
    \end{centering}
\end{figure}
\begin{table}[t]
\caption{Results}
\label{dice}
\begin{center}
\begin{tabular}{llllll}
\multicolumn{1}{c}{\bf ARCHITECTURE}  &\multicolumn{1}{c}{\bf IS}&\multicolumn{1}{c}{\bf N}&\multicolumn{1}{c}{\bf MF}&\multicolumn{1}{c}{\bf DICE}
\\ \hline \\
Unet96X2048X4	&96	&4	&2048	&0.81\\
Unet96X1024X4	&96	&4	&1024	&0.73\\
Unet96X512X4	&96	&4	&512	&0.61\\
Unet96X256X4	&96	&4	&256	&0.32\\
Unet192X1024X5	&192	&5	&1024	&0.54\\
Unet96X1024X5	&96	&5	&1024	&0.66\\
Unet48X1024X4	&48	&4	&1024	&0.31\\
Unet96X1024X4-SE	&96	&4	&1024	&0.75\\
Unet96X512X4-SE	&96	&4	&512	&0.65\\
Unet96X256X4-SE	&96	&4	&256	&0.35\\
\end{tabular}
\end{center}
\end{table}
Fig. \ref{fig:seg} shows an example of application of this architecture to our case with minor modifications. We were able to train the model for the segmentation task with less than 400 labelled images (with the accuracy on the test set ~89\%).

\section{Results and discussion}
\label{results}
Dice similarity coefficient (Dice score) is often used to quantify how closely the results of the model match hand annotated ground truth segmentation. We used continuous versions of the Dice score \citep{dice}. \\
Based on the results of our experiments, we can conclude that combination of state-of-the-art techniques can improve the performance of the U-net on satellite data. Increasing the number of layers leads to a small increase in the model accuracy, so as the increasing the number of filters in the last convolutional layers. Some increase could be also achieved by adding residual connections, but the best accuracy was achieved by addition of SE blocks. These results are displayed in Table.\ref{dice}, where "IS" stands for the "Input Size", "N" stands for the "Number of downsample/upsample stages" and "MF" stands for the "Max filters on lower block". \\
To conclude, this research showed how AI and earth observation data can be deployed to tackle the SDGs. Further, we sketched a case for the different roles of AI from mapping and mediating data for SDGs. The future work programme could expand the roles to adjudicating and ranking by building an evaluative infrastructure that allows to develop generally accepted and effectively implemented methodologies for SDG indicators. The ‘systems thinking’ approach we propose here, brings three benefits. First, it makes the impact of public and private investors in relation to the SDGs in a geographical area. This allows, secondly, for countries to be evaluated and ranked according to their performance, with further reputational effects. Finally, for developers and researchers this means that AI can be more effectively applied to solutions directly contributing towards specific SDGs. 


\appendix
\begin{table}[t]
\caption{Role of AI and EO in addressing SDGs}
\label{role-addressing}
\begin{center}
\begin{tabular}{llll}
\multicolumn{2}{c}{\bf SDGs TARGETS \& INDICATORS} & \multicolumn{1}{c}{\bf EXPLANATION} & \multicolumn{1}{c}{\bf TIER}
\\ \hline \\
\multicolumn{4}{c}{\bf  1st generation AI-EO application}
\\ \\
6. Clean water 	&6.1.1*	&Change in the extent of water-	&III\\
\& sanitation && related ecosystems over time&\\

&6.3.1	&Proportion of wastewater safely treated	&III\\
&6.3.2	&Proportion of bodies of water with good 	&III\\
&&ambient water quality&\\
&6.6.1	&Change in the extent of water-related 	&III\\
&	&ecosystems over time	&\\
9. Industries, innovation	&9.1.1	&Proportion rural population 	&III\\
\& infrastructure & & living within 2km of all-season road&\\
11. Sustainable cities	&11.3.1*	&Ratio of land consumption rate and 	&II\\
\& communities&&population growth rate&\\
&11.7.1	&Average proportion of the built surface of &II\\
&	& the cities corresponding to open spaces &\\
&	& for the public use of all	&\\
15. Life on land	&15.1.1	&Forest area as a proportion of total 	&I\\
	&	& land area	&\\
&15.2.1	&Progress towards sustainable forest 	&III\\
&	& management	&\\
&15.3.1*	&Proportion of land that is degraded over 	&III\\
&	& total land area	&\\
&15.4.2	&Mountain Green Cover Index	&III\\ \\
\multicolumn{4}{c}{\bf  2nd generation AI-EO application}
\\ \\
1. No poverty 	&1.2.2*&	Proportion of men, women and children 	&II\\
&&	of all ages living in poverty in all its 	&\\
&&	dimensions according to national definitions	&\\
&1.4.1	&Proportion of population living	&III\\
&	&in households with access to basic services	&\\
2. Zero hunger	&2.4.1*	&Proportion of agricultural area under 	&III\\
	&	& productive and sustainable agriculture 	&\\
&2.5.1*	&Number of plant and animal genetic resources&	III\\
&	& for food and agriculture secured in either  &	\\
&	& medium- or long-term conservation facilities&	\\
6. Clean water 
	&6.5.2	&Proportion of transboundary basin area 	&III\\
\& sanitation	&	& with an operational arrangement for water  	&\\
	&	& cooperation 	&\\
11. Sustainable cities	&11.2.1 	&Proportion of population that has convenient 	&II\\
\& communities	& 	&access to public transport, by sex, age 	& \\
	& 	& and persons with disabilities 	& \\
13. Climate action	&13.1.2	&Number of countries that adopt and	&II\\
	&	&implement national disaster risk reduction &\\
	&	& strategies in line with the Sendai 	&\\
	&	& Framework	&\\
	&13.1.3	&Proportion of local governments that adopt 	&II\\
	&	& and implement local disaster risk reduction  	&\\
	&	& strategies in line with national strategies	&\\
	&13.2*	&Integrate climate change measures into &	I\\
	& 	&national policies, strategies and planning&	 \\
14. Life below water&	14.1.1	&Index of coastal eutrophication and &	III\\
&& floating plastic debris density&	III\\
\end{tabular}
\end{center}
\end{table}
\bibliography{iclr2019_conference}
\bibliographystyle{iclr2019_conference}

\end{document}